%% file: main.tex
\documentclass[10pt,twocolumn,letterpaper]{article}

\usepackage[pagenumbers]{cvpr} 

\input{preamble}

\usepackage{times}
\usepackage{epsfig}
\usepackage{graphicx}
\usepackage{amsmath}
\usepackage{amssymb}
\usepackage{xcolor}
\usepackage{multirow}
\usepackage{adjustbox}
\usepackage{float}
\usepackage{pifont}
\usepackage{svg}
\usepackage{algorithm}
\usepackage{algpseudocode}
\usepackage{tikz}
\usepackage{booktabs}
\usepackage{subcaption}
\usepackage{svg}

\definecolor{blue}{RGB}{34,34,225}
\definecolor{myGreen}{RGB}{34, 139, 34}
\definecolor{red}{RGB}{225, 34, 34}

\definecolor{cvprblue}{rgb}{0.21,0.49,0.74}
\usepackage[pagebackref,breaklinks,colorlinks,citecolor=cvprblue]{hyperref}

\def\paperID{307} 
\def\confName{3DV\xspace}
\def\confYear{2026\xspace}

\title{
Avatar4D: Synthesizing Domain-Specific 4D Humans\\ for Real-World Pose Estimation
}

\author{\parbox{16cm}{\centering
    {\large Jerrin Bright$^{1,3}$, Zhibo Wang$^{1,2,3}$, Dmytro Klepachevskyi\textsuperscript{1,2,3}, Yuhao Chen$^{1,3}$, Sirisha Rambhatla$^{2,3}$, David Clausi$^{1,3}$ }, John Zelek$^{1,3}$\\
    {\normalsize
    $^1$ Vision and Image Processing Lab,  
    $^2$ Critical ML Lab,
    $^3$ University of Waterloo, Canada\\
    {\tt\small {\{jerrin.bright, zhibo.wang, dklepachevskyi, yuhao.chen1, sirisha.rambhatla, \\
    dclausi, jzelek\}}@uwaterloo.ca}
    }
}}

\newcommand{\proposed}{\texttt{Avatar4D}}
\newcommand{\dataset}{\texttt{Syn2Sport}}
\begin{document}
\maketitle

\input{sec/0_abstract}    
\input{sec/1_intro}
\input{sec/2_lit}
\input{sec/3_method}
\input{sec/4_dataset}
\input{sec/5_exp}
\input{sec/6_conc}

\newpage

{
    \small

\input{main.bbl}
}

\end{document}

%% file: preamble.tex
\usepackage[dvipsnames]{xcolor}


%% file: sec/0_abstract.tex
\begin{abstract}
We present {\proposed}, a real-world transferable pipeline for generating customizable synthetic human motion datasets tailored to domain-specific applications. Unlike prior works, which focus on general, everyday motions and offer limited flexibility, our approach provides fine-grained control over body pose, appearance, camera viewpoint, and environmental context, without requiring any manual annotations. To validate the impact of {\proposed}, we focus on sports, where domain-specific human actions and movement patterns pose unique challenges for motion understanding. In this setting, we introduce {\dataset}, a large-scale synthetic dataset spanning sports, including baseball and ice hockey. {\dataset} features high-fidelity 4D (3D geometry over time) human motion sequences with varying player appearances rendered in diverse environments. We benchmark several state-of-the-art pose estimation models on {\dataset} and demonstrate their effectiveness for supervised learning, zero-shot transfer to real-world data, and generalization across sports. Furthermore, we evaluate how closely the generated synthetic data aligns with real-world datasets in feature space. Our results highlight the potential of such systems to generate scalable, controllable, and transferable human datasets for diverse domain-specific tasks without relying on domain-specific real data.
\end{abstract}

%% file: sec/1_intro.tex
\section{Introduction}
\label{sec:intro}


Understanding human motion is central to a wide range of vision tasks, from activity recognition and biomechanics to sports analytics \cite{bright2024pitchernet, soccernet} and interaction modeling \cite{PROX, sahmr}. General-purpose human pose datasets \cite{coco, mpii, 3dpw, h36m} provide broad coverage of everyday activities, but they lack domain-specific motions and rarely include challenging poses or sport-specific dynamics. Existing sports-focused datasets \cite{lsp, soccernet, fsjump3d, tokenclipose, mitigatingblur} offer more targeted coverage but remain limited to popular sports such as soccer, leaving many activities and their unique motions underrepresented, such as skating in ice hockey or pitching in baseball. This gap motivates the need for domain-specific large-scale datasets that reflect their diversity and complexity.

Synthetic data offers a promising alternative, providing scalable access to diverse and precisely labeled samples. While recent synthetic datasets \cite{agora, surreal, bedlam, pumarola20193dpeople, gta} have enabled progress in modeling general everyday human activities, they do not capture domain-specific challenges such as high acceleration, unusual joint configuration, or unusual temporal dynamics. In addition, they lack fine-grained control over body posture, camera viewing perspective, appearance variability (such as clothing, body shape, and textures), lighting, and background context, which we refer to as the \textit{configurable simulation parameters}.

This raises a key challenge: \textit{How can we generate synthetic human datasets that are configurable to domain requirements and effective enough to substitute for real-world data in those domains?}

\begin{figure*}[t]
  \centering
  \includegraphics[width=0.95\linewidth]{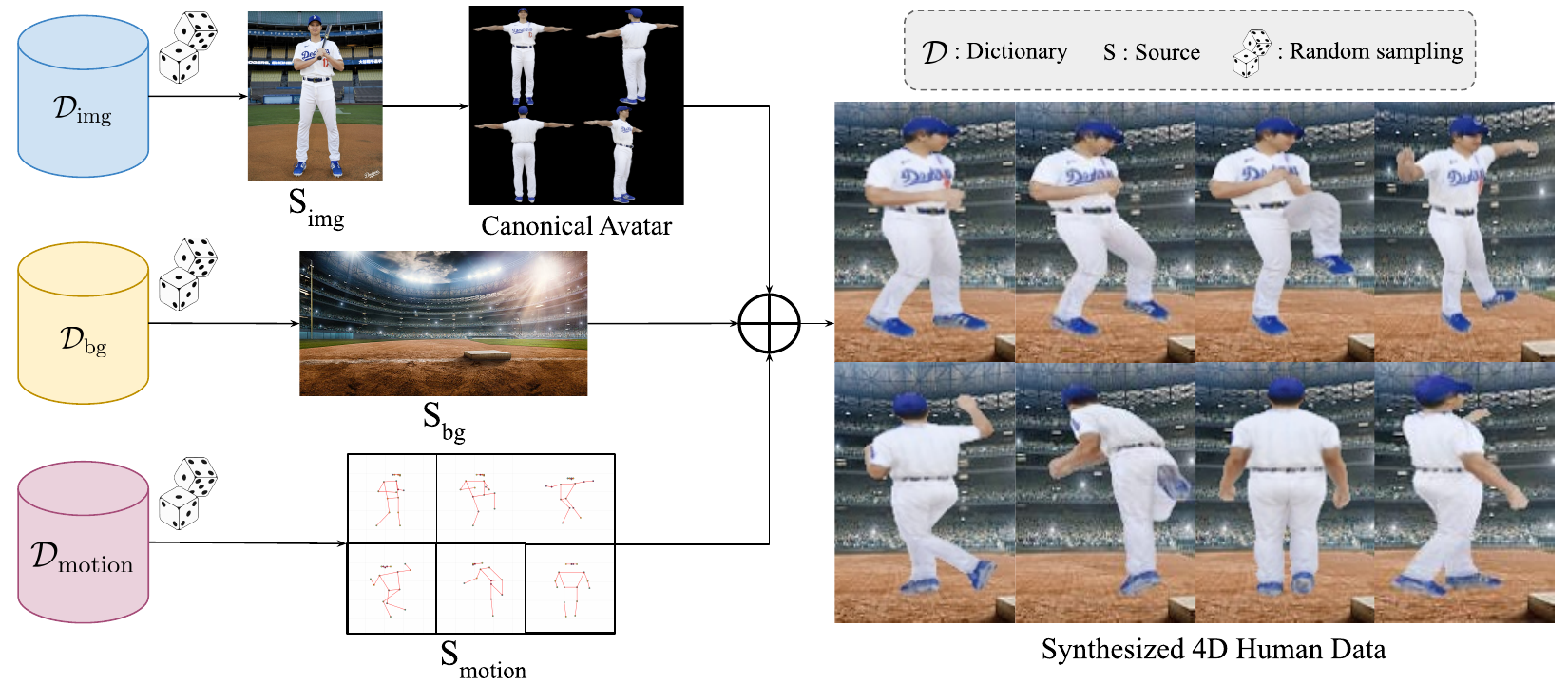}
  \vspace{-10px}
   \caption{\textbf{Overview of Avatar4D architecture.} We sample a random source image ($\text{S}_\text{img}$), source background ($\text{S}_\text{bg}$) and motion sequence ($\text{S}_\text{motion}$) from three dictionaries: $\mathcal{D}_\text{img}$ for domain-relevant person images, $\mathcal{D}_\text{bg}$ for scene/background images, and $\mathcal{D}_\text{motion}$ for motion sequences. These inputs are then combined to generate domain-specific synthetic 4D human data.}
   \label{fig:overview}
\end{figure*}

Motivated by this question, we introduce {\proposed}, a generation pipeline for synthetic human motion datasets tailored to specific domains. The key challenge is to capture diverse, unbiased distributions that still transfer well to real data. As illustrated in Figure \ref{fig:overview}, our pipeline operates in three stages: (1) we build a dictionary of domain-relevant inputs, including source person images, motion sequences from internet videos, and background scenes; (2) each person image is lifted into a canonical 3D asset, which is then animated by deforming it according to the chosen motion sequence; and (3) the animated 3D human is rendered against the corresponding background to form a realistic frame. The final output is a large-scale collection of high-fidelity 4D human sequences (spanning diverse identities, motions, and environments) that can be \textit{directly leveraged for training and evaluation in real-world conditions.} For example, {\proposed} can generate basketball players dribbling on a court, construction workers moving at a site, or pedestrians navigating streets, enabling domain-specific benchmarks for human-centric tasks such as pose estimation, 3D human modeling, and motion segmentation. 


To demonstrate and validate the capabilities of {\proposed} in generating realistic, domain-specific synthetic data, we use it to create {\dataset}, a large-scale synthetic dataset spanning sports domains, including baseball and ice hockey. We specifically focused on these sports because of their distinct motion patterns: fine-grained single-athlete actions in baseball and fast-paced skating action in ice hockey. {\dataset} contains high-fidelity human motion sequences rendered across diverse environments with rich appearance variations, providing a controlled testbed to evaluate the effectiveness of the proposed {\proposed} approach. Importantly, to address our central research question, we conducted extensive experiments using the {\dataset} dataset, including zero-shot synthetic-to-real pose transfer, domain generalizability, and feature-space alignment.
The results demonstrate that {\proposed} enables the generation of synthetic datasets that are both configurable to domain-specific requirements and transferable to real data, thereby reducing reliance on real-world labels.
Our contributions can be summarized as follows:

\begin{enumerate}
    \item We propose {\proposed}, a real-world transferable synthetic pipeline for generating domain-specific datasets, with precise control over the configurable simulation parameters.
    \item We introduce {\dataset}, a large-scale synthetic dataset featuring high-fidelity 4D human motion across sports domains (baseball and icehockey).
    \item We conduct extensive experiments on zero-shot domain generalization, synthetic-to-real transfer across different sports, and feature-space alignment.
\end{enumerate}

%% file: sec/2_lit.tex
\section{Related Works}
\label{sec:lit}

\subsection{Real Datasets}

Real-world datasets have been foundational in advancing Human Pose Estimation (HPE). HPE aims to extract the 2D human pose of a person given a single image. Some commonly benchmarked datasets include COCO \cite{coco} and MPII \cite{mpii}, which provide large-scale annotated images of everyday activities, supporting robust keypoint detection. There are some other task-specific datasets, such as sports-focused datasets like LSP \cite{lsp} offer annotated images for athletic movements, while datasets such as PoseTrack \cite{posetrack}, OCHuman \cite{ochuman}, and CrowdPose \cite{li2019crowdpose} capture multi-person interactions with heavy occlusions.

While these datasets have driven significant progress in 2D HPE, they exhibit several limitations: (1) Most datasets focus on everyday activities, offering limited coverage of specialized domains such as sports, dance, or rehabilitation tasks; (2) Rare or complex poses are underrepresented, which restricts model generalization; and (3) Capturing and labeling 2D pose data is time-consuming and are prone to human annotation bias. These limitations with existing real-world datasets make it necessary to develop alternative approaches to collect domain-specific training data without having to manually label it.

\begin{figure*}[t]
  \centering
  \includegraphics[width=0.95\linewidth]{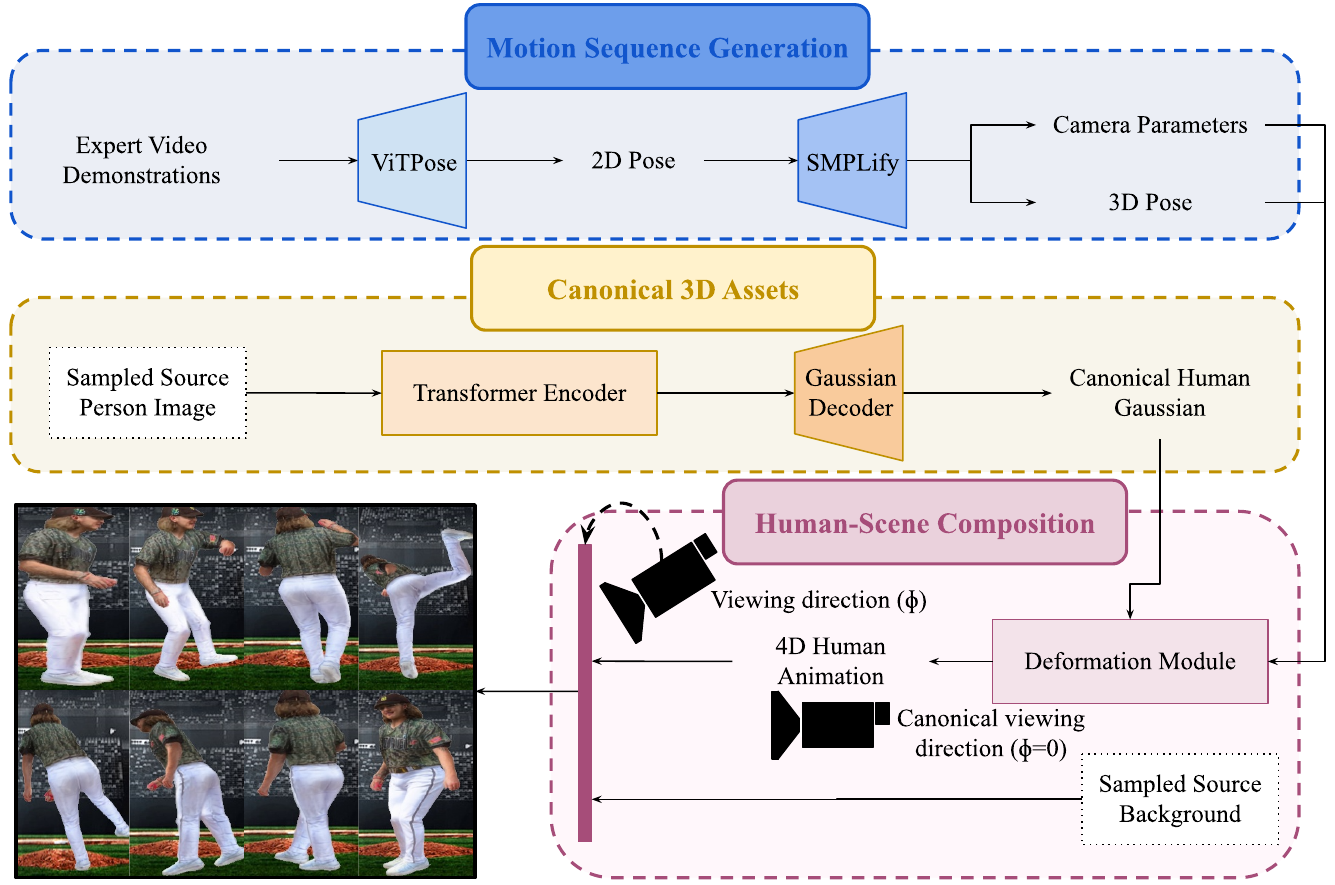}
  \vspace{-10px}
   \caption{\textbf{Architecture of Avatar4D.} The proposed pipeline consists of three key stages. First, a motion sequence is constructed from expert demonstrations collected from online sources, producing 3D poses along with corresponding camera parameters. Next, canonical 3D human assets are generated from sampled source person images. Finally, in the human-scene composition stage, these assets are deformed according to the 3D poses and rendered against sampled source background. }
   \label{fig:arch}
\end{figure*}

\subsection{Synthetic Datasets}

Synthetic datasets provide a scalable and flexible alternative by leveraging 3D modeling, motion capture, and rendering techniques. Notable examples include SURREAL \cite{surreal}, AGORA \cite{agora}, BEDLAM \cite{bedlam}, and 3DPeople \cite{pumarola20193dpeople}, which provide annotated 2D and 3D poses with synthetic images. More recently, methods such as Gen4D \cite{gen4d} allow dynamic 4D human synthesis with fine-grained control over body posture, appearance, and background; however, it doesn't transfer well to real-world scenarios.

However, existing synthetic datasets still have important limitations: (1) Most synthetic datasets emphasize everyday actions, lacking domain-specific coverage for tasks like sports analytics, rehabilitation, or industrial applications; (2) Artifacts introduced by rendering pipelines or background generation can reduce realism, impeding direct transfer to real-world models; and (3) Restricted fine-grained control over pose sequences, camera viewpoints, or environmental context simultaneously, which is crucial for generating task-specific training data.

These gaps highlight the need for a domain-adaptive, high-fidelity synthetic pipeline that can generate realistic 4D human motion data for target domains while ensuring transferability to real-world scenarios.




%% file: sec/3_method.tex
\section{Methodology}
\label{sec:method}

\subsection{Overview}

The goal of the proposed {\proposed} generation pipeline is to synthesize a large-scale 4D-human sequence with controllable motion, realistic geometry, and photorealistic rendering. To achieve this, we decouple the problem into three stages: motion generation, asset creation, and sequence rendering. So we begin with controllable motion data generation, which is then paired with canonical 3D asset generation from monocular human images. The generated motion is used to deform the 3D asset, and the resulting sequence is rasterized against a static background. Figure \ref{fig:arch} illustrates the overall pipeline, while subsequent sections detail each component of the pipeline.

\subsection{Motion Sequence Generation} \label{sec:motion}

Prior works \cite{surreal, bedlam} typically utilize motion sequences derived from MoCap datasets \cite{cmu_mocap, amass}, which are commonly used to drive the deformation of 3D human assets. However, these datasets are often limited in diversity and fail to capture the breadth of motion found in in-the-wild, domain-specific scenarios, such as those encountered in sports or street activities. To address this limitation, we construct a dictionary of motion sequences that better reflects real-world variability. Unlike recent generative methods that synthesize motion from text prompts using motion diffusion models \cite{mdm, flowmdm, priormdm}, we curate motion data directly from internet videos of target activities, allowing us to ground the motion in real-world dynamics.

Specifically, we extract whole-body 2D keypoints (pseudo) from these videos using a pretrained pose estimator \cite{vitpose}. These pseudo keypoints are then lifted to 3D body parameters by fitting an SMPL-X mesh using SMPLify-X \cite{smplx}. While the extracted pseudo keypoints may be noisy, they are only used to drive synthetic avatar generation. This ensures that the final generated avatars still provide perfect image–pose pairs under controllable camera viewpoints. The resulting sequences serve as pseudo-ground truth to guide the 3D assets for downstream composition.

\subsection{Canonical 3D Assets}

3D asset generation has traditionally relied on CGI pipelines and animation experts. This manual process is time-consuming and inherently limits variability, especially in task-specific settings where asset appearance (e.g., clothing, body shape, accessories) needs to adapt to the domain. A promising alternative is the use of 3D Gaussian Splatting (3DGS) \cite{3dgs}, which allows for fast and high-fidelity rendering of dynamic or clothed human avatars without explicit mesh reconstruction.

Recent advances in 3DGS for humans have explored asset creation from various input modalities, ranging from multi-view images \cite{gauhuman, 3dgs-avatar, gart}, text-guided templates \cite{humangaussian, clothedreamer, laga}, to single-view images \cite{lhm, idol}. In our pipeline, we adopt the LHM \cite{lhm} method, which enables reconstruction of animatable 3D avatars from a single input image. Compared to multi-view reconstruction methods \cite{gauhuman, 3dgs-avatar}, which require cumbersome data collection and calibration, or text-prompt-driven methods \cite{humangaussian, laga}, which often generate avatars with limited realism and domain fidelity, LHM offers a practical solution for scalable dataset generation. Each avatar is represented as a collection of 3D Gaussians, where each Gaussian is parameterized by a centroid $p \in \mathbb{R}^3$, an anisotropic scale $\sigma \in \mathbb{R}^3$, and a quaternion $r \in \mathbb{R}^4$ that encodes its rotation. Additionally, an opacity value $\rho \in [0, 1]$ controls the transparency of the Gaussian, while a feature vector $f \in \mathbb{R}^C$, encoded using spherical harmonics, captures view-dependent appearances. These parameters collectively form the tuple $\chi = (p, r, f, \rho, \sigma)$, which defines the full set of properties for each point in the 3DGS representation. 

\subsection{Human-Scene Composition}

\subsubsection{View Space Rendering}

Given the canonical 3DGS asset parameters ($\chi$), i.e., the canonical (undeformed rest-pose) representation of the avatar in its reference space, we first transform the canonical avatar to target view space using Linear Blend Skinning (LBS) based on the motion sequence. Each Gaussian ($\mathbf{x}_i$) is transformed as shown in Equation \eqref{eq:lbs}. 

\begin{equation}
    \mathbf{x}_{i'} = \sum_{j=1}^{J} w_{ij} \mathbf{T}_j \mathbf{x}_i
    \label{eq:lbs}
\end{equation}

\noindent where $w_{ij}$ are the skinning weights and $\mathbf{T}_j \in \mathbb{R}^{4\times4}$ are the per-joint transformation matrices computed via forward kinematics. The deformed Gaussians are then rendered through differentiable splatting under the target camera parameters $\pi_t$ to produce the RGB image $\hat{I}$ and alpha mask $\hat{M}$. The final composite image $\mathcal{I}$ is obtained by alpha-blending the rasterized foreground with the background image $I_{\text{bg}}$:

$$\mathcal{I} = \hat{I} \odot \mathbf{\hat{M}} + {I}_{\text{bg}} \odot (1-\mathbf{\hat{M}})$$

\noindent where ${I}_{\text{bg}}$ is the background image, which is alpha-blended with the rasterized foreground. 

\noindent \textbf{Keypoints Transformation.} The 2D keypoints generated from Section \ref{sec:motion} must be transformed to align spatially with the rendered view space. This transformation includes three sequential steps: centering concerning the original camera principal point, scaling to match the canonical render resolution $r$, and repositioning within the final image frame. Formally, given an original keypoint $(x, y)$ and its associated principal point $(c_x^{\text{old}}, c_y^{\text{old}})$, the transformed keypoint coordinates $(x', y')$ in the rendered image space of dimensions $(w_{\text{new}}, h_{\text{new}})$ are computed as:

\begin{align}
x' &= (x - c_x^{\text{old}}) \cdot \frac{r}{w_{\text{new}}} + \frac{w_{\text{new}}}{2} \\
y' &= (y - c_y^{\text{old}}) \cdot \frac{r}{h_{\text{new}}} + \frac{h_{\text{new}}}{2}
\end{align}

%% file: sec/4_dataset.tex
\begin{figure}[t]
  \centering
  \includegraphics[width=\linewidth]{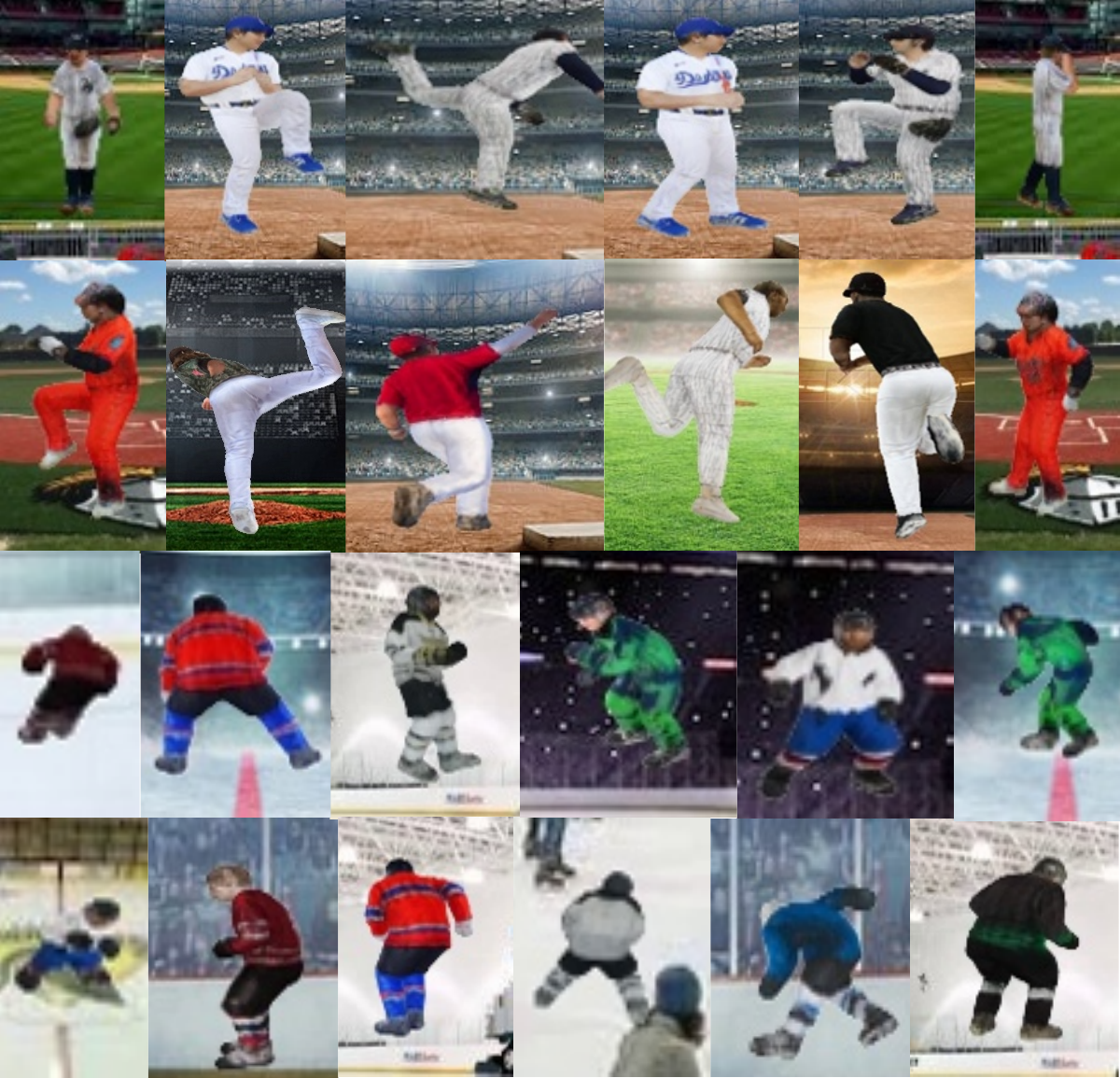}
  \vspace{-20px}
  \caption{\textbf{Sample images from the Syn2Sport dataset.} All samples were generated using the proposed {\proposed} pipeline.}
   \label{fig:sampledata}
\end{figure}

\section{Dataset}
\label{sec:dataset}

We present {\dataset}, a large-scale synthetic human motion dataset covering sports such as baseball and ice hockey, generated using the {\proposed} pipeline. {\dataset} comprises 21,203 clips across all splits, totaling roughly 4.48 million frames and over 2,400 minutes of play time after augmentation. Some sample images from our dataset are illustrated in Figure \ref{fig:sampledata}.

The dataset is split into training and validation sets per sport, as summarized in Table \ref{tab:sportspal-data-split}. Each sample includes 2D pose annotations in COCO-WB \cite{cocowb} format, 3D pose annotations, and full 3D human pose and shape parameters represented in the SMPL-X model. This combination of 2D and 3D annotations enables training and evaluation across a broad range of human pose estimation tasks, from 2D keypoint detection to detailed 3D human mesh recovery.

\noindent \paragraph{Augmentations.} To enhance viewpoint diversity in {\dataset}, we implement stochastic camera sampling for each motion sequence. Specifically, the camera’s extrinsic parameters (3D position relative to the subject and its orientation), defined by elevation and azimuth angles, are randomly sampled per sequence. This produces varying perspectives capturing the human from different angles.

%% file: sec/5_exp.tex
\section{Experiments}
\label{sec:exp}

\subsection{Evaluation}

\noindent \textbf{Datasets.} We evaluate the performance of the models trained with the data generated from {\dataset} with two real-world datasets. MLBPitchDB \cite{mitigatingblur}, which is a dataset focusing on baseball players containing high motion blur and self-occlusion. For ice hockey, we use a large-scale dataset \cite{tokenclipose} which is annotated from NHL broadcast videos. Furthermore, for zero-shot experimentation across other sports domains, we use two real-world datasets: a soccer dataset \cite{soccer} and a lacrosse dataset \cite{tokenclipose}. To provide a baseline comparison against models trained on large-scale real-world data, we also use COCO-WholeBody (COCO-WB) \cite{cocowb}, an extension of COCO \cite{coco} with 133 keypoints covering the face, body, hands, and feet. For our experiments, we specifically use the first 23 keypoints corresponding to the body and feet. Additionally, we compare against SportPAL \cite{gen4d}, a synthetic sports benchmark that most closely aligns with our setting by reconstructing controllable human motions within contextual backgrounds.
 
\noindent \textbf{Metrics.} To evaluate pose estimation performance, we report the Average Precision (AP) at multiple thresholds. AP measures the percentage of predicted keypoints falling within a specified distance (in pixels) from the ground truth, normalized by the subject’s bounding box size. For most experiments, we use thresholds of 5, 10, and 15 pixels. For the more challenging zero-shot sports transfer setting, we adopt slightly relaxed thresholds of 15, 20, and 25 pixels to account for the increased domain gap.

To measure the similarity between the real and synthetic data distributions, we computed the Fréchet Inception Distance (FID) \cite{fid} and Kernel Inception Distance (KID) \cite{kid} between the deep feature embeddings of the synthetic and real data. FID measures the distance between the Gaussian approximations of the feature distributions, while KID computes an unbiased squared Maximum Mean Discrepancy (MMD) between the distributions.

\begin{table}[t]
\centering
\caption{\small \textbf{Syn2Sport Data Split.} Data statistics across the baseball and ice hockey sports.}
\vspace{-10px}
\adjustbox{width=\linewidth}{
\setlength{\tabcolsep}{6pt}
\begin{tabular}{lccccc}
\toprule
\textbf{Sport} & \textbf{Split} & \#\textbf{Subjects} & \#\textbf{Clips} & \#\textbf{Frames}  & \begin{tabular}{c}\textbf{Play Time}\\(mins)\end{tabular} \\ 
\midrule
\multirow{2}{*}{Baseball} 
    & Train  & 15 & 1,017 & 585,000 & 319\\
    & Valid  & 5 & 254 & 146,413 & 87 \\
\midrule
\multirow{2}{*}{Ice hockey} 
    & Train  & 15 & 15,946 & 3,002,000 & 1,608 \\
    & Valid  & 5 & 3,986 & 750,569 & 477\\
\midrule
\textbf{Total} & - & \textbf{40} & \textbf{21,203} & \textbf{4,483,982} & \textbf{2,491} \\ 
\bottomrule
\end{tabular}
}
\label{tab:sportspal-data-split}
\end{table}

\begin{table}[t]
\caption{\small {\textbf{Benchmarking Syn2Sport on 2D pose estimation models.} The best result is highlighted in \textbf{bold} format.}}
\vspace{-10px}
\centering
\adjustbox{width=\linewidth}{
\setlength{\tabcolsep}{5pt}
\begin{tabular}{lcccccc}
\toprule
\multirow{2}{*}{Method} 
& \multicolumn{3}{c}{Baseball} 
& \multicolumn{3}{c}{Icehockey} \\
\cmidrule(lr){2-4}
\cmidrule(lr){5-7}
& AP$^5\textcolor{red}{\uparrow}$ & AP$^{10}\textcolor{red}{\uparrow}$ & AP$^{15}\textcolor{red}{\uparrow}$ 
& AP$^5\textcolor{red}{\uparrow}$ & AP$^{10}\textcolor{red}{\uparrow}$ & AP$^{15}\textcolor{red}{\uparrow}$ \\
\midrule
DETR \cite{detr} & 74.5 & 78.1 & 82.2 & 31.7 & 41.4 & 48.6 \\
HRNet \cite{hrnet} & 83.6 & 83.6 & 96.0 & 43.0 & 64.4 & 74.1 \\
UDP \cite{udp} & 85.2 & 88.5 & 92.6 & 38.9 & 61.9 & 75.3 \\
ViTPose \cite{vitpose} & 83.6 & 95.7 & 97.9 & \textbf{55.3} & \textbf{82.4} & \textbf{91.0} \\
TokenPose~\cite{tokenhmr} & \textbf{89.1} & \textbf{97.5} & \textbf{98.7} & 54.2 & 81.1 & 90.4 \\

\bottomrule
\end{tabular}
}
\label{tab:exp:main}
\end{table}

\subsection{Main Experiments}

\textbf{Benchmarking on Syn2Sport Dataset.} Since {\dataset} provides large-scale synthetic sports sequences with ground truth 2D pose annotations, we use it as a benchmark to assess the quality of synthetic data for HPE. Specifically, Table \ref{tab:exp:main} evaluates state-of-the-art 2D HPE models on the baseball and icehockey subsets using keypoint accuracy (AP), thereby quantifying how well these models perform when applied to these domains.

\noindent \textbf{Zero-Shot Synthetic-to-Real Transfer.} The ultimate goal of the proposed method and its resulting dataset is to generate synthetic data that closely represents the real domain. To evaluate this, we analyze how models trained on the synthetic data transfer to real-world datasets. We consider three training configurations: training on the COCO-WB dataset alone, training on the synthetic dataset alone, and training on both datasets combined. Performance is measured on two real-world sports datasets: baseball \cite{mitigatingblur} and ice hockey \cite{tokenclipose}, with all models trained using TokenPose \cite{tokenpose}.

\begin{figure}[t]
  \centering
  \includegraphics[width=\linewidth]{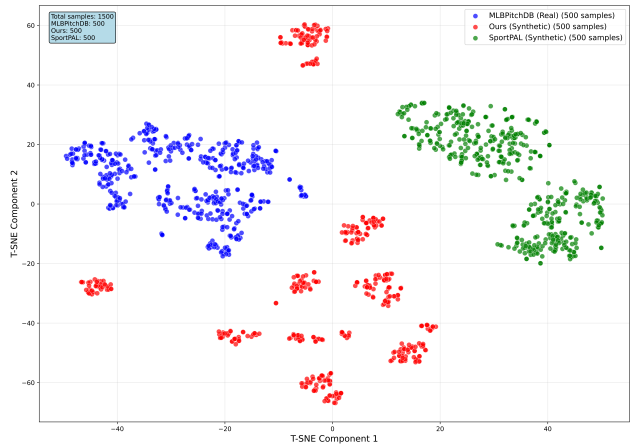}
  \caption{\textbf{t-SNE visualization of feature embeddings.} Real samples (red) and synthetic samples (blue) show substantial overlap, indicating strong similarity between the datasets.}
  \label{fig:tsne}
\end{figure}

The results, as shown in Table \ref{tab:zero_shot_2d_combined} demonstrate that incorporating synthetic data significantly improves performance. Specifically, adding synthetic data yields 55.96\% and 114.5\% improvements in AP$^5$ on the real-world baseball and ice hockey datasets, respectively, compared to training on COCO-WB alone. While COCO-WB contains some real baseball examples and slightly outperforms the synthetic-only model on MLBPitchDB, combining COCO-WB with synthetic data produces the best results. In contrast, COCO-WB alone fails to generalize to ice hockey due to the absence of skating-related samples, where the synthetic-only model achieves substantial improvements. These results demonstrate that \textit{the synthetic data generated from {\proposed} can effectively substitute for real-world, domain-specific training samples}, enabling models to be pretrained on general-purpose datasets and then finetuned solely on our synthetic samples for robust results.

\begin{figure*}
  \centering
  \includegraphics[width=0.95\linewidth]{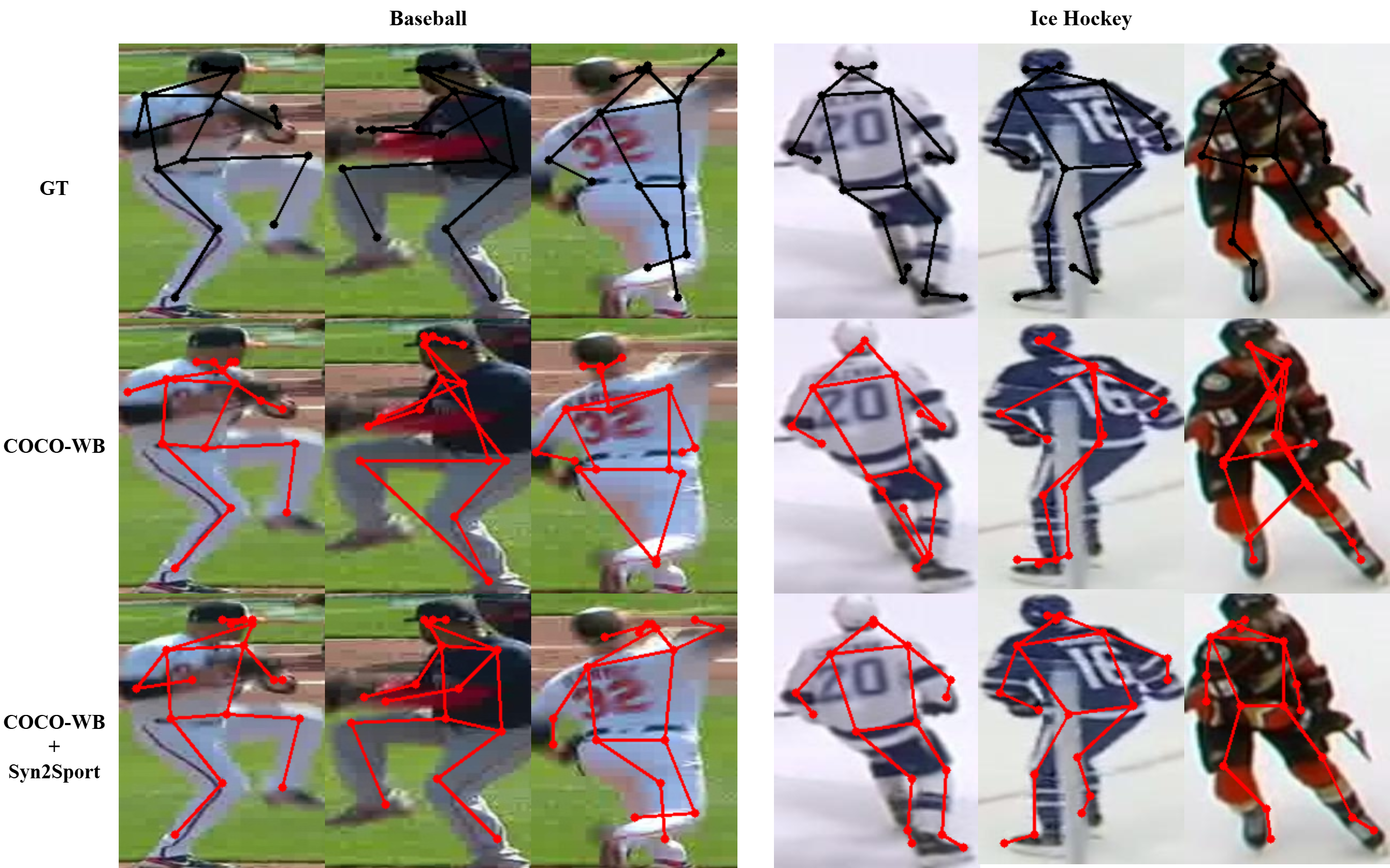}
  \vspace{-10px}
   \caption{\textbf{Comparison of 2D Pose Transfer Results.} Models are trained on the synthetic subset of {\dataset} and the real-world COCO-WB dataset, and evaluated on real-world baseball \cite{mitigatingblur} and ice hockey \cite{tokenclipose} datasets. COCO-WB+{\dataset} denotes pretraining on COCO-WB followed by finetuning on our synthetic subset, and GT denotes the ground truth from the respective dataset.}
   \label{fig:transfer}
\end{figure*}

Figure \ref{fig:transfer} shows examples of pose transfer from our synthetic dataset to real-world images. Models trained on {\dataset} accurately predict keypoints on real-world baseball and ice hockey datasets, demonstrating improved alignment compared to training solely on COCO-WB.

\noindent \textbf{Evaluation of Synthetic Data Fidelity.} To quantitatively and qualitatively assess how closely our synthetic dataset resembles the real data, we conducted a feature-space similarity analysis. First, we extract high-level image features \cite{vgg} from both the synthetic and real datasets to capture semantic and perceptual image information. Results as shown in Table \ref{tab:fid_comparison} demonstrate that our synthetic dataset improves FID by 34\% and KID by 54\%, demonstrating better feature-space alignment compared to SportPAL \cite{gen4d}.

To complement the quantitative metric, we performed t-SNE dimensionality reduction on the same feature embeddings, projecting them into a 2D space as shown in Figure \ref{fig:tsne}. t-SNE was computed on feature embeddings with perplexity of 30 and 1000 iterations. In the resulting visualization, real and synthetic samples were collected from MLBPitchDB, SportPAL, and {\dataset}, confirming that our synthetic samples were positioned closer to real data than the data generated from SportPAL, indicating improved semantic and perceptual similarity.

\begin{figure}[t]
  \centering
  \includegraphics[width=\linewidth]{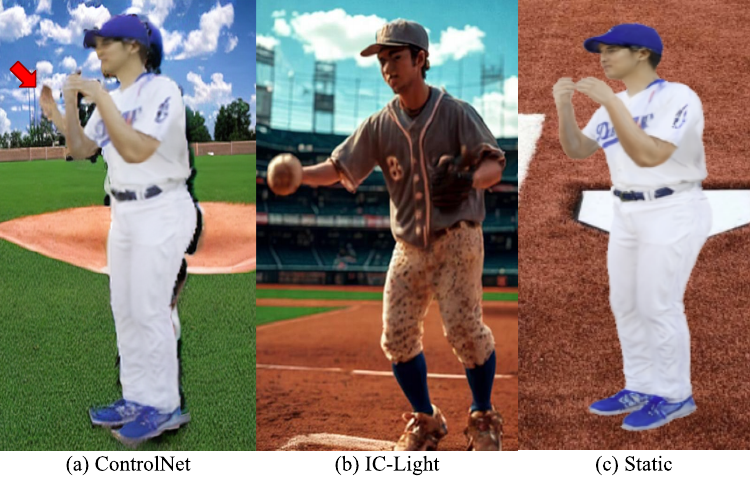}
  \vspace{-20px}
  \caption{\textbf{Background generation using different methods.} Samples generated using ICLight, ControlNet, and static backgrounds, illustrating their impact on the realism of the synthetic human data.}
  \label{fig:bg}
\end{figure}

\begin{figure*}
  \centering
  \includegraphics[width=0.95\linewidth]{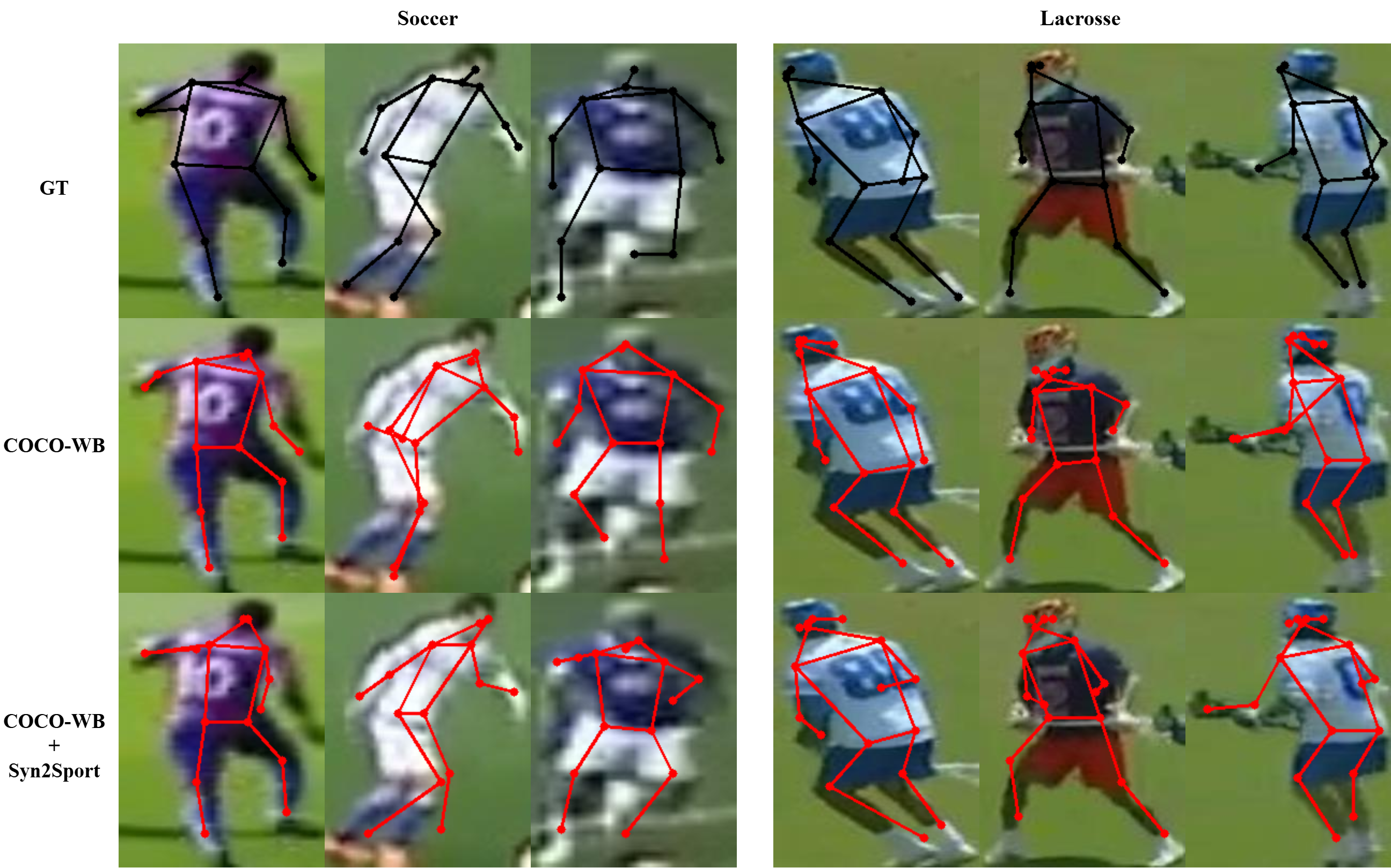}
  \vspace{-10px}
  \caption{\textbf{Comparison of Zero-Shot 2D Pose Transfer Results.} Models are trained on the synthetic baseball and ice hockey subsets of {\dataset} and evaluated on the real-world soccer \cite{soccer} and lacrosse \cite{tokenclipose} datasets, respectively.}
   \label{fig:zero-shot}
\end{figure*}

\subsection{Ablation Study}

\textbf{Impact of Background on Synthetic-to-Real Transfer.} We evaluate the impact of different background generation methods on synthetic-to-real transfer for pose estimation, comparing static image backgrounds, IC-Light \cite{iclight}, and ControlNet \cite{zhang2023adding}. Figure \ref{fig:bg} shows examples of backgrounds generated using each approach. Generative methods often alter the human asset itself: IC-Light produces an animation-like, artificial appearance, while ControlNet occasionally introduces spurious artifacts, such as an extra hand (highlighted by the red arrow), which degrade model performance. In contrast, static backgrounds preserve the fidelity of the human model, achieving superior performance as confirmed by the quantitative analysis in Table \ref{tab:background_2d}.

\begin{table}[t]
\caption{\small \textbf{2D Pose Transfer Performance} on real-world baseball \cite{mitigatingblur} and ice hockey \cite{tokenclipose} datasets.}
\vspace{-10px}
\centering
\adjustbox{width=\linewidth}{
\setlength{\tabcolsep}{6pt}
\begin{tabular}{lccccccccc}
\toprule
\multirow{2}{*}{Training Data} 
& \multicolumn{3}{c}{Baseball} 
& \multicolumn{3}{c}{Ice hockey} \\
\cmidrule(lr){2-4} \cmidrule(lr){5-7}
& AP$^5\textcolor{red}{\uparrow}$ & AP$^{10}\textcolor{red}{\uparrow}$ & AP$^{15}\textcolor{red}{\uparrow}$ 
& AP$^5\textcolor{red}{\uparrow}$ & AP$^{10}\textcolor{red}{\uparrow}$ & AP$^{15}\textcolor{red}{\uparrow}$ \\
\midrule
COCO-WB & 29.3 & 56.0 & 71.3 & 13.1 & 39.5 & 58.3 \\
{\dataset} & 23.0 & 41.3 & 55.0 & 19.5 & 44.9 & 62.3 \\
COCO-WB+{\dataset} & \textbf{45.7} & \textbf{77.6} & \textbf{90.7} & \textbf{28.1} & \textbf{61.2} & \textbf{82.0} \\
\bottomrule
\end{tabular}
}
\label{tab:zero_shot_2d_combined}
\end{table}

\begin{table}[t]
\centering
\caption{\textbf{FID comparison between synthetic datasets and real data.} Lower values indicate closer alignment with the real data.}
\vspace{-10px}
\begin{tabular}{l cc}
\hline
Dataset & FID $\textcolor{blue}{\downarrow}$ & KID $\textcolor{blue}{\downarrow}$\\
\hline
Real Data (reference) & 0 & 0 \\ 
SportPAL & 21.6 & 0.09 \\ 
{\dataset} (Ours) & \textbf{14.3} & \textbf{0.04}\\ 
\hline
\end{tabular}
\label{tab:fid_comparison}
\end{table}

\noindent \textbf{Cross-Sport Zero-Shot Generalization.} To evaluate domain generalization, we train pose estimation models on synthetic data from a single sport (e.g., baseball or ice hockey) and test them zero-shot on real-world datasets from different but related sports (e.g., soccer and lacrosse, respectively). While some motion patterns or equipment may be shared, such as stick handling in ice hockey and lacrosse, these sports differ in visual context, movement dynamics, and scene composition. This setup allows us to assess whether models trained on synthetic data from one domain can transfer effectively to related domains, highlighting the robustness and versatility of our synthetic dataset.

\begin{table}[t]
\caption{\small \textbf{Impact of Background on 2D Pose Transfer} on the real-world baseball \cite{mitigatingblur} and ice hockey \cite{tokenclipose} datasets.}
\vspace{-10px}
\centering
\adjustbox{width=\linewidth}{
\setlength{\tabcolsep}{6pt}
\begin{tabular}{lccccccccc}
\toprule
\multirow{2}{*}{Background} 
& \multicolumn{3}{c}{Baseball} 
& \multicolumn{3}{c}{Ice hockey} \\
\cmidrule(lr){2-4} \cmidrule(lr){5-7}
& AP$^5\textcolor{red}{\uparrow}$ & AP$^{10}\textcolor{red}{\uparrow}$ & AP$^{15}\textcolor{red}{\uparrow}$ 
& AP$^5\textcolor{red}{\uparrow}$ & AP$^{10}\textcolor{red}{\uparrow}$ & AP$^{15}\textcolor{red}{\uparrow}$ \\
\midrule
ControlNet \cite{zhang2023adding} & 38.1 & 67.7 & 84.0 & 18.4 & 43.4 & 71.9 \\
IC-Light \cite{iclight} & 3.4 & 11.3 & 20.2 & 1.3 & 6.7 & 15.4 \\
Static & \textbf{45.7} & \textbf{77.6} & \textbf{90.7} & \textbf{21.7} & \textbf{52.2} & \textbf{74.8} \\
\bottomrule
\end{tabular}
}
\label{tab:background_2d}
\end{table}

\begin{table}[t]
\caption{\small \textbf{Zero-Shot Generalization Across Sports Domains} with data trained on {\dataset} dataset.}
\vspace{-10px}
\centering
\adjustbox{width=\linewidth}{
\setlength{\tabcolsep}{8pt}
\begin{tabular}{lcccc}
\toprule
Train & Test (Real) & AP$^{15}\textcolor{red}{\uparrow}$ & AP$^{20}\textcolor{red}{\uparrow}$ & AP$^{25}\textcolor{red}{\uparrow}$ \\
\midrule
COCO-WB (Real) & Lacrosse \cite{tokenclipose} & 64.7 & 72.2 & 81.6 \\
Ice hockey (Synthetic) & Lacrosse \cite{tokenclipose} & 66.5 & 76.8 & 83.3 \\
COCO-WB (Real) & Soccer \cite{soccer} & 71.3 & 78.2 & 83.6 \\
Baseball (Synthetic) & Soccer \cite{soccer} & 67.1 & 76.1 & 82.4 \\
\bottomrule
\end{tabular}
}
\label{tab:zero_shot_sport}
\end{table}

Table \ref{tab:zero_shot_sport} demonstrates strong zero-shot performance, highlighting the model’s ability to generalize across both visual and kinematic variations using only synthetic supervision. Notably, the TokenPose model trained on our dataset outperforms on the real-world lacrosse test set and shows comparable performance on the soccer dataset relative to the model trained on the manually annotated COCO-WB dataset. These results underscore the advantage of our synthetic pipeline, which achieves competitive or superior performance without relying on any manual annotations.

%% file: sec/6_conc.tex
\section{Discussion}

Our findings provide several practical insights for designing synthetic human motion datasets. First, static backgrounds outperform context-aware generative methods for human-centric motion tasks, as generative backgrounds can introduce an artificial look in the human model that reduces transferability to real data. Second, synthetic data substantially improves real-world performance when combined with models pretrained on COCO, achieving over 50\% AP gains on baseball and ice hockey benchmarks. Finally, zero-shot cross-domain generalization is possible, with models trained on one sport transferring to related sports by leveraging shared motion and appearance patterns. Overall, these results highlight that prioritizing human asset fidelity and leveraging synthetic augmentation enables effective, transferable, and task-flexible datasets.

Importantly, the proposed pipeline establishes a domain-adaptive workflow for pose estimation: given a new target domain, a model can be initialized with weights from a pretrained general-purpose dataset (e.g., COCO) and subsequently \textit{fine-tuned exclusively on synthetic data generated from our pipeline}. This eliminates the need for domain-specific real-world annotations while retaining high performance, enabling scalable, task-specific, and transferable human motion modeling.

\section{Conclusion}

We present {\proposed}, a real-world transferable pipeline for generating customizable synthetic human motion datasets for domain-specific applications. Our approach provides fine-grained control over pose, appearance, camera, and environment, enabling the synthesis of high-fidelity 4D human data tailored to target tasks. Using this system, we release {\dataset}, a large-scale synthetic dataset spanning multiple sports, and demonstrate that models trained on it achieve strong supervised performance, zero-shot real-world transfer, and cross-sport generalization. Feature-space analysis confirms that our synthetic data closely aligns with real-world distributions. These results highlight the potential of controllable synthetic data pipelines to reduce reliance on real annotations while enabling scalable and transferable human motion modeling.


\label{sec:conc}